\documentclass{article}

\usepackage[final]{neurips_2019}

\usepackage[utf8]{inputenc} % allow utf-8 input
\usepackage[T1]{fontenc}    % use 8-bit T1 fonts
\usepackage{hyperref}       % hyperlinks
\usepackage{url}            % simple URL typesetting
\usepackage{booktabs}       % professional-quality tables
\usepackage{amsfonts}       % blackboard math symbols
\usepackage{nicefrac}       % compact symbols for 1/2, etc.
\usepackage{microtype}      % microtypography

\usepackage{enumitem}

\usepackage{algorithm}
\usepackage{algorithmic}
\usepackage{booktabs}

\usepackage{wrapfig}
\usepackage{changepage}
\usepackage{amsmath}
\usepackage{amsthm}
\usepackage{amssymb}
\usepackage{amsfonts,eucal,amsbsy}
\usepackage{mathtools}
\usepackage{multirow}
\usepackage[normalem]{ulem}
\usepackage{slashbox}
\usepackage{xcolor}

\usepackage{caption}
\usepackage{subcaption}

\DeclareMathOperator*{\argmax}{arg\,max}
\DeclareMathOperator*{\argmin}{arg\,min}
\newcommand{\ignore}[1]{}
\newcommand{\vect}[1]{\mathbf{#1}}

\newcommand{\struct}[1]{\boldsymbol{#1}}

\newcommand*\arc{{\fontfamily{pbk}\fontseries{db}\selectfont+}{\fontfamily{pbk}\fontseries{db}\selectfont+}}

\title{Episodic Memory in Lifelong Language Learning}

\author{%
  Cyprien de Masson d'Autume, Sebastian Ruder, Lingpeng Kong, Dani Yogatama\\
  DeepMind\\
  London, United Kingdom \\
  \texttt{\{cyprien,ruder,lingpenk,dyogatama\}@google.com} \\
}

\begin{document}

\maketitle

\begin{abstract}
We introduce a lifelong language learning setup where a model
needs to learn from a stream of text examples without any 
dataset identifier.
We propose an episodic memory model
that performs sparse experience replay and 
local adaptation to mitigate catastrophic forgetting in this setup.
Experiments on text classification and question answering
demonstrate the complementary benefits of sparse experience replay
and local adaptation to allow the model 
to continuously learn from new datasets.
We also show that the space complexity of
the episodic memory module can be 
reduced significantly ($\sim$50-90\%)
by randomly choosing which examples to store
in memory with a minimal decrease in performance.
We consider an episodic memory component as a 
crucial building block of 
general linguistic intelligence and see our model
as a first step in that direction.
\end{abstract}

%We also explore a method to
%reduce memory capacity using the concept of forgettable example.

\section{Introduction}
The ability to continuously learn and accumulate knowledge 
throughout a lifetime and reuse it effectively to adapt to a
new problem quickly is a hallmark of general intelligence.
State-of-the-art machine learning models work well
on a single dataset given enough training examples, but
they often fail to isolate and reuse 
previously acquired knowledge
when the data distribution shifts (e.g., 
when presented with a new dataset)---a 
phenomenon known as \emph{catastrophic forgetting} \citep{mccloskey1989catastrophic,ratcliff1990connectionist}.

The three main approaches to address catastrophic 
forgetting are 
based on: (i) augmenting the loss
function that is being minimized during training
with extra terms (e.g., a regularization term, 
an optimization constraint)
to prevent model parameters learned on a new dataset 
from significantly deviating from parameters
learned on previously seen datasets \citep{ewc, zenke, chaudhry},
(ii) adding extra learning phases such as a knowledge distillation phase, an experience replay \citep{progresscompress, experiencereplay},
and (iii) augmenting the 
model with an episodic memory module \citep{mbpa}.
Recent methods have shown that these 
approaches can be combined---e.g.,
by defining optimization constraints using samples from the
episodic memory \citep{gem, agem}.

In language learning, progress in unsupervised pretraining 
\citep{elmo,ulmfit,bert} has driven advances in
many language understanding tasks \citep{kitaev,lee}.
However, these models have been shown to require a lot of
in-domain training examples, rapidly overfit to particular datasets,
and are prone to catastrophic forgetting \citep{glipaper},
making them unsuitable as a model of general linguistic intelligence.

In this paper, we investigate the role of 
episodic memory for learning a model of language
in a lifelong setup. 
We propose to use such a component
for \emph{sparse experience replay} and \emph{local adaptation}
to allow the model to continually learn from examples drawn from 
different data distributions.
In experience replay, we randomly select
examples from memory to retrain on.
Our model only performs experience replay
very sparsely to consolidate newly acquired knowledge 
with existing knowledge in the memory into the model.
We show that a 1\% experience replay to learning 
new examples ratio is sufficient.
Such a process bears some similarity 
to memory consolidation in human learning \citep{McGaugh248}.
In local adaptation, we follow
Memory-based Parameter 
Adaptation \citep[MbPA;][]{mbpa} and
use examples retrieved from memory to update model
parameters used to make a prediction of a particular
test example.

Our setup is different 
from a typical lifelong learning setup.
We assume that the model 
only makes one pass over the training examples, similar to \citet{agem}.
However, we also assume neither our 
training nor test examples 
have dataset identifying information (e.g., 
a dataset identity, a dataset descriptor).
We argue that our lifelong language learning setup---where
a model is presented with 
a stream of examples without an explicit 
identifier about which dataset (distribution)
the examples come from---is a 
realistic setup to learn a general linguistic intelligence model.\footnote{Contrast this with a more common setup 
where the model learns in a multitask setup \citep{mtl,decanlp}.}
Our experiments focus on lifelong language 
learning on two tasks---text 
classification and question answering.\footnote{\citet{decanlp} show that many language processing 
tasks (e.g., classification, summarization, natural language 
inference, etc.) can 
be formulated as a question answering problem.}
% We argue that our lifelong language learning setup---where a
% model is presented with question-answer examples
% without an explicit identifier about which dataset (distribution)
% the examples come from---is a more realistic setup
% to learn a general linguistic intelligence model.

Our main contributions in this paper are:
\begin{itemize}
\item We introduce a lifelong language learning setup where the model
needs to learn from a stream of examples from many datasets 
(presented sequentially) in one pass, and no dataset boundary
or dataset identity is given to the model.
\item We present an episodic memory model (\S{\ref{sec:model}}) 
that augments an encoder-decoder model with a memory module.
Our memory is a key-value memory that stores
previously seen examples for sparse experience replay and local adaptation.
\item We leverage progress in unsupervised pretraining
to obtain good memory key representations and discuss
strategies to manage the space complexity of the memory module.
%We explore methods based on random sampling, 
%the concept of forgettable examples \citep{forgetexamples},
%and adaptive posterior sampling \citep{adaptivesampling}. We 
%show that random sampling is a strong baseline.
\item We compare our proposed method to baseline and state-of-the-art continual learning methods and demonstrate its efficacy on text classification and question answering tasks (\S{\ref{sec:experiments}}). 
\end{itemize}
%\dycomment{add rapid adaptation, if results are positive on this}

%\cite{Bapna2019} is a related recent approach for NMT. They use n-gram based overlap based on Transformer subword embeddings instead of sentence similarity for retrieval.

\section{Model} 
\label{sec:model}
\begin{figure}
    \centering
    \includegraphics[scale=0.33]{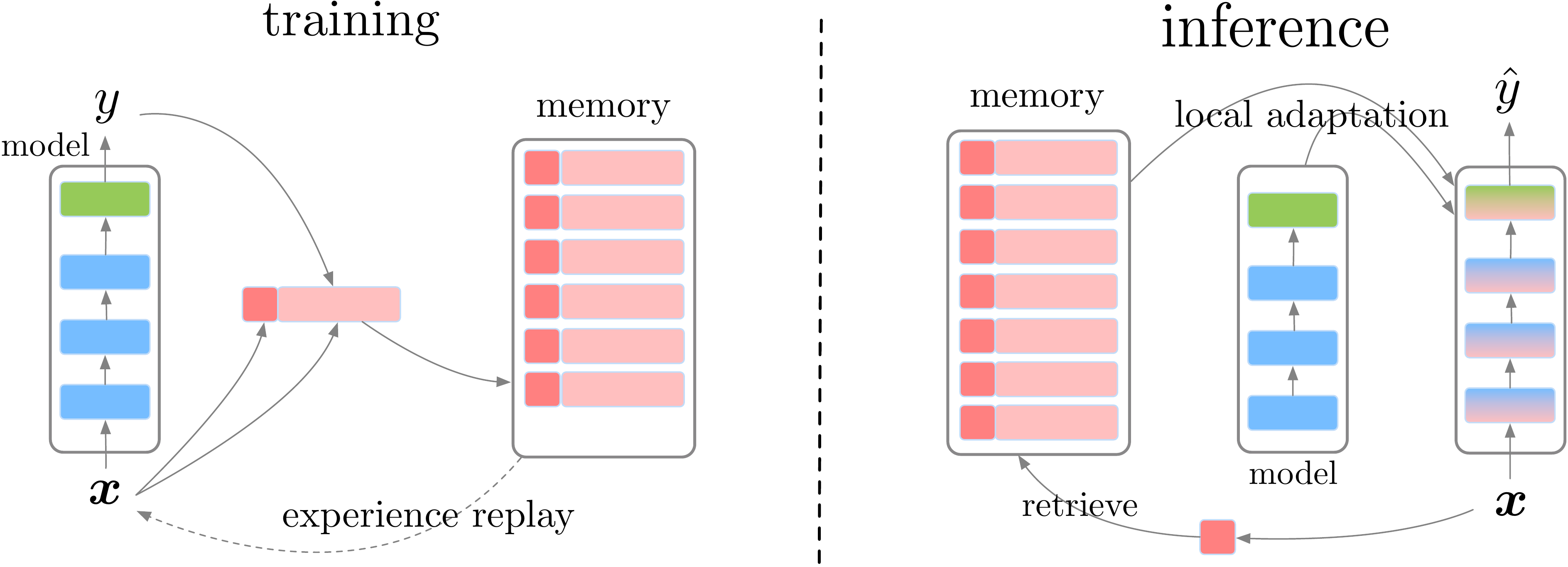}
    \caption{An illustration of our model and how it interacts with the key-value memory module during training (left) and inference (right). During training, newly seen examples are used to update the base model and stored in the memory. At certain intervals, we sample examples from the memory and perform gradient updates on the base model (experience replay). During inference, we retrieve examples whose keys are similar to a test example under consideration to fine-tune the model (local adaptation). We use the fine-tuned model to make a prediction and then discard it---keeping the base model for other predictions.}
    \label{fig:model}
\vspace{-0.5cm}
\end{figure}

We consider a continual (lifelong) learning setup where a model 
needs to learn from a stream of 
training examples $\{\struct{x}_t, y_{t}\}_{t=1}^T$.
We assume that all our training examples in the series 
come from multiple datasets of the same task 
(e.g., a text classification task, a question answering task),
and each dataset comes one after the other.
Since all examples come from the same task,
the same model can be 
used to make predictions on all examples.
A crucial difference between our continual learning setup
and previous work is that we do not assume
that each example comes with a dataset descriptor (e.g., a dataset identity).
As a result, the model does not know which dataset
an example comes from and when a dataset boundary has been crossed during training.
The goal of learning is to find parameters $\vect{W}$ that minimize the negative log probability of training examples under our model:
\begin{align*}
\mathcal{L}(\vect{W}) = - \sum_{t=1}^T \log p(y_t \mid \struct{x}_t; \vect{W}).
\end{align*}

Our model consists of three main components: (i) an example encoder, (ii) a
task decoder, and 
(iii) an episodic memory module.
Figure~\ref{fig:model} shows an illustration of our complete model.
We describe each component in detail in the following.

\subsection{Example Encoder}
Our encoder is based on the Transformer architecture \citep{transformer}.
We use the state-of-the-art text encoder \textsc{BERT} \citep{bert} 
to encode our input $\struct{x}_t$. \textsc{BERT}
is a large Transformer
pretrained on a large unlabeled corpus on two unsupervised tasks---masked
language modeling and next sentence prediction.
Other architectures such as 
recurrent neural networks or convolutional 
neural networks can also be used as the example encoder.

In text classification, 
$\struct{x}_t$ is a document to be classified;
\textsc{BERT} produces a vector representation of each
token in $\struct{x}_t$, which includes a special 
beginning-of-document symbol \texttt{CLS} as $x_{t,0}$.
In question answering,
$\struct{x}_t$ is a concatenation of a context 
paragraph $\struct{x}^{\text{context}}_t$ and a 
question $\struct{x}^{\text{question}}_t$ separated by a 
special separator symbol \texttt{SEP}.

\subsection{Task Decoder}
%Since all examples in a series are from the same task,
%our model uses one only decoder to make predictions.
In text classification, following the original \textsc{BERT}
model, we take the representation
of the first token $x_{t,0}$ from \textsc{BERT} (i.e., the special
beginning-of-document symbol) and 
add a linear transformation and a softmax layer to predict the class
of $\struct{x}_t$.
\begin{align*}
p(y_t = c\mid \struct{x}_t) = \frac{\exp(\vect{w}_c^{\top}\vect{x}_{t,0})}{\sum_{y \in \mathcal{Y}} \exp(\vect{w}_y^{\top}\vect{x}_{t,0})}
\end{align*}
Note that since there is no dataset descriptor
provided to our model,
this decoder is used to predict all classes in all datasets, which we assume 
to be known in advance.

For question answering, our decoder predicts an 
answer span---the start and end indices of the correct answer 
in the context. Denote the length
of the context paragraph by $M$, and $\struct{x}^{\text{context}}_t = \{x^{\text{context}}_{t,0}, \ldots, x^{\text{context}}_{t,M}\}$.
Denote the encoded representation of the $m$-th token in the context
by $\vect{x}^{\text{context}}_{t,m}$.
Our decoder has two sets of parameters: $\vect{w}_{\text{start}}$
and $\vect{w}_{\text{end}}$.
The probability of each context token being the start of the answer
is computed as:
\begin{align*}
p(\texttt{start} = x^{\text{context}}_{t,m} \mid \struct{x}_t) = \frac{\exp(\vect{w}_{\text{start}}^{\top}\vect{x}^{\text{context}}_{t,m})}{\sum_{n=0}^M \exp(\vect{w}_{\text{start}}^{\top}\vect{x}^{\text{context}}_{t,n})}.
\end{align*}
We compute the probability of the end index of the answer analogously using $\vect{w}_{\text{end}}$.
The predicted answer is the span with the highest probability
after multiplying the start and end probabilities. 
We take into account that the start index of an answer 
needs to precede its end index by setting the probabilities of invalid spans to zero.

\subsection{Episodic Memory}
Our model is augmented with an episodic memory module that
stores previously seen examples throughout its lifetime.
The episodic memory module is used for sparse experience replay
and local adaptation to prevent catastrophic forgetting
and encourage positive transfer.
We first describe the architecture of our episodic memory module,
before discussing how it is used at training
and inference (prediction) time in \S{\ref{sec:training}}.

The module is a key-value memory block.
We obtain the key representation of $\struct{x}_t$ (denoted by $\vect{u}_t$) 
using a key network---which is a pretrained \textsc{BERT} model separate from the
example encoder. We freeze the key network to prevent key representations from drifting as data distribution changes (i.e. the problem that the key of 
a test example tends to be closer to keys of recently stored examples).

For text classification, our key is
an encoded representation of the first token of 
the document to be classified,
so $\vect{u}_t = \vect{x}_{t,0}$ (i.e., the special beginning-of-document symbol).
For question answering, we first take the question part of the input 
$\struct{x}_t^{\text{question}}$. We encode it using the key network
and take the first token as the key vector
$\vect{u}_t = x_{t,0}^{\text{question}}$.\footnote{Our 
preliminary experiments suggest that using only 
the question as the key
slightly outperforms using the full input.
Intuitively, given a
question such as ``Where was Barack Obama from?''
and an article about Barack Obama, we would like to retrieve examples
with similar questions rather
than examples with articles about the same topic,
which would be selected if we used the entire input (question and context)
as the key.}
For both tasks, we store the input and the label
$\langle \struct{x}_t, y_t \rangle$ 
as its associated memory value.

\paragraph{Write.}
If we assume that the model has unlimited
capacity, we can write all training examples into the memory.
However, this assumption is unrealistic in practice.
We explore a simple writing strategy that relaxes this constraint
based on random write.
%We explore three additional writing strategies
%that relax this constraint: (i) random write, (ii) forgettable examples,
%and (iii) adaptive posterior sampling.
In random write, we randomly decide whether to write a newly seen
example into the memory with some probability.
We find that this is a strong baseline that outperforms
other simple methods based on 
surprisal \citep{adaptivesampling}
and the concept of forgettable examples \citep{forgetexamples} in
our preliminary experiments.%\footnote{Other work that
%uses an episodic memory per task is not directly applicable in the
%setting without task descriptors.}
We leave investigations of more sophisticated selection methods to
future work.
%In forget examples, we use the concept of forgettable examples
%\citep{forgetexamples} and design a method that removes (forgets)
%examples from the memory if they are correctly predicted ...
%Lastly, we also investigate a method based on 
%adaptive posterior sampling \citep{adaptivesampling} ...
%\dycomment{TODO: sebastian add methods here}

\paragraph{Read.}
Our memory has two retrieval mechanisms: (i) random sampling
and (ii) $K$-nearest neighbors. We use random sampling
to perform sparse experience replay and $K$-nearest neighbors
for local adaptation, which are described 
in \S{\ref{sec:training}} below.

\section{Training and Inference}
\label{sec:training}
Algorithm~\ref{alg:training} and Algorithm~\ref{alg:inference} 
outline our overall training and inference procedures.

\paragraph{Sparse experience replay.}
At a certain interval throughout the learning period, 
we uniformly sample from stored examples in the memory
and perform gradient updates of the encoder-decoder network
based on the retrieved examples.
Allowing the model to perform 
experience replay at every timestep would 
transform the problem of continual learning
into multitask learning.
While such a method will protect the
model from catastrophic forgetting, it is expensive
and defeats the purpose of a lifelong learning setup.
Our experience replay procedure is designed to 
be performed very sparsely. 
In practice, we randomly retrieve 100 examples 
every 10,000 new examples. Note that similar to the base training
procedure, we only perform one gradient update 
for the 100 retrieved examples.

\paragraph{Local adaptation.}
At inference time, given a test example, 
we use the key
network to obtain a query vector of the test example
and query the memory to retrieve $K$ nearest 
neighbors using the Euclidean distance function.
We use these $K$ examples to perform local adaptation,
similar to Memory-based Parameter Adaptation \citep{mbpa}.
Denote the $K$ examples retrieved for the $i$-th test example
by $\{\struct{x}_{i}^k, y_{i}^k\}_{k=1}^K$.
We perform gradient-based local adaptation
to update parameters of the encoder-decoder model---denoted 
by $\vect{W}$---to obtain local parameters $\vect{W}_i$ to be used
for the current prediction as follows:\footnote{
Future work can explore cheaper alternatives to gradient-based updates for
local adaptation (e.g., a Hebbian update
similar to the update that is 
used in plastic networks; Miconi et al., 2018). \nocite{hebbian}}
\begin{align}
\label{eq:local}
\vect{W}_i = \argmin_{\vect{\tilde{W}}} \lambda \Vert \vect{\tilde{W}} - \vect{W}\Vert_2^2 - \sum_{k=1}^K \alpha_k \log p(y_{i}^k \mid \struct{x}_i^k; \vect{\tilde{W}}),
\end{align}
where $\lambda$ is a hyperparameter, $\alpha_k$ is the weight
of the $k$-th retrieved example and $\sum_{k=1}^K \alpha_k = 1$.
In our experiments, we assume that all $K$ 
retrieved examples are equally important regardless
of their distance to the query vector and set $\alpha_k = \frac{1}{K}$.
Intuitively, the above procedure locally adapts parameters 
of the encoder-decoder network to be better at predicting
retrieved examples from the memory (as defined by having a 
higher probability of predicting $y_i^k$), while keeping it close to the
base parameters $\vect{W}$.
Note that $\vect{W}_i$ is only used to make a prediction 
for the $i$-th example, and the parameters are 
reset to $\vect{W}$ afterwards.
In practice, we only perform $L$ local adaptation gradient steps
instead of finding the true minimum of Eq.~\ref{eq:local}.

\begin{minipage}[t]{.48\textwidth}
\vspace{-0.5cm}
\begin{algorithm}[H]%\small
\caption{Training}
\label{alg:training}
\begin{algorithmic}
   \STATE {\bfseries Input:} training examples $\langle\struct{x}_t, y_t\rangle_{t=1}^T$, replay interval $R$
   \STATE {\bfseries Output:} parameters $\vect{W}$, memory $\struct{M}$
   \FOR{$t=1$ {\bfseries to} $T$}
   \IF{$t \text{ mod } R = 0$}
   \STATE Sample $S$ examples from $\struct{M}$.
   \STATE Perform gradient updates on $\vect{W}$. \COMMENT{experience replay}
   \ENDIF
   \STATE Receive a training example $\langle\struct{x}_t, y_t\rangle$.
   \STATE Perform a gradient update on $\vect{W}$ to minimize $- \log p(y_t \mid \struct{x}_t; \vect{W})$.
   \IF{store example}
   \STATE Write $\langle\struct{x}_t, y_t\rangle$ to memory $\struct{M}$.
   \ENDIF
   \ENDFOR
\end{algorithmic}
\end{algorithm}
\end{minipage}
\hfill
\begin{minipage}[t]{.48\textwidth}
\vspace{-0.5cm}
\begin{algorithm}[H]%\small
\caption{Inference}
\label{alg:inference}
\begin{algorithmic}
   \STATE {\bfseries Input:} test example $\struct{x}_i$, parameters $\vect{W}$, memory $\struct{M}$
   \STATE {\bfseries Output:} test prediction $\hat{y}_i$
   \STATE Compute query representation $\vect{u}_i$ from $\struct{x}_i$.
   \STATE Find $K$ nearest neighbors of $\vect{u}_i$ from $\struct{M}$.
   \STATE $\vect{W}_i \leftarrow \vect{W}$
   \FOR{$l=1$ {\bfseries to} $L$}
   \STATE Perform a gradient update on $\vect{W}_i$ to minimize Eq. \ref{eq:local}. \COMMENT{local adaptation}
   \ENDFOR
   \STATE $\hat{y}_i = \argmax_y p(y \mid \struct{x}_i; \vect{W}_i)$
\end{algorithmic}
\end{algorithm}
\end{minipage}

\section{Experiments}
\label{sec:experiments}
In this section, we evaluate our proposed model
against several baselines on text classification and question answering tasks.

\subsection{Datasets}
%Table~\ref{tbl:datasets} shows a summary of datasets
%used in our experiments. We describe each dataset in more
%details below.

%\begin{table}[h]
%\caption{Descriptive statistics of datasets used in our experiments.}
%    \label{tbl:setup}
%    \centering
%    \begin{tabular}{|c|l|c|r|r|r|}
%     \hline
%      \textbf{Task} & \textbf{Dataset} & \textbf{Genre} & \textbf{\# of train} & \textbf{\# of test} & \textbf{\# of class} \\
%      \hline
%       \multirow{5}{*}{classification} & AG News & news & 50,000 & 5,000 & 4 \\
%       & DBPedia &  & 50,000 & 5,000 & 14 \\
%       & Amazon & reviews & 50,000 & 5,000 & 5 \\
%       & Yahoo & & 50,000 & 5,000 & 10 \\
%       & Yelp & reviews & 50,000 & 5,000 & 5 \\
%       \hline
%       \multirow{4}{*}{question answering} & SQuAD && 500,000 & 5,000 & \\
%       & TriviaQA---Web && 500,000 & 5,000 & \\
%       & TriviaQA---Wikipedia && 500,000 & 5,000 & \\
%       & QuAC && 500,000 & 5,000 & \\
%      \hline
%    \end{tabular}
%\end{table}

\paragraph{Text classification.}
We use publicly available text classification datasets from \citet{zhangcnn} to evaluate our models (\footnotesize\url{http://goo.gl/JyCnZq}\normalsize).
This collection of datasets 
includes text classification datasets from diverse domains such as
news classification (AGNews), sentiment analysis (Yelp, Amazon), 
Wikipedia article classification (DBPedia), and questions and answers categorization (Yahoo).
Specifically, we use AGNews (4 classes), Yelp (5 classes), DBPedia (14 classes), Amazon (5 classes), and Yahoo (10 classes) datasets.
Since classes for Yelp and Amazon datasets have similar 
semantics (product ratings), we merge the classes for these two datasets.
In total, we have 33 classes in our experiments.
These datasets have varying sizes. For example, AGNews 
is ten times smaller than Yahoo. We create a balanced version
all datasets used in our experiments by randomly 
sampling 115,000 training examples
and 7,600 test examples from all 
datasets (i.e.,  the size of the smallest training and test
sets). We leave investigations of 
lifelong learning in unbalanced datasets to future work.
In total, we have 575,000 training examples and
38,000 test examples.%\footnote{For replicability, 
%we release our sampled training and test
%sets at \url{http://anonymous.com}.}

\paragraph{Question answering.}
We use three question answering 
datasets: SQuAD 1.1 \citep{squad}, TriviaQA \citep{triviaqa}, and QuAC \citep{quac}.
These datasets have different characteristics. 
SQuAD is a reading comprehension dataset constructed from
Wikipedia articles.
It includes almost 90,000 training examples 
and 10,000 validation examples.
TriviaQA is a dataset with question-answer pairs written 
by trivia enthusiasts and evidence collected retrospectively 
from Wikipedia and the Web.
There are two sections of TriviaQA, Web and Wikipedia, which we treat as separate datasets.
The Web section contains 76,000 
training examples and 10,000 (unverified) validation examples,
whereas the Wikipedia section has about 60,000 training examples
and 8,000 validation examples.
QuAC is an information-seeking dialog-style dataset where 
a student asks questions about a Wikipedia article 
and a teacher answers with a short excerpt from the article.
It has 80,000 training examples and approximately 7,000 validation
examples.

\subsection{Models}
We compare the following models in our experiments:
\begin{itemize}
\item \textsc{Enc-Dec}: a standard encoder-decoder model without any episodic memory module.
\item \textsc{A-GEM} \citep{agem}: Average Gradient Episodic Memory model that defines constraints on the gradients 
that are used to update model parameters based on retrieved examples from the memory. In its original formulation, \textsc{A-GEM} requires dataset identifiers
and randomly samples examples from previous datasets. 
We generalize it to the setting without dataset
identities by randomly sampling from the episodic
memory module at fixed intervals, similar to our method.
\item \textsc{Replay}: a model that uses stored examples 
for sparse experience replay without local adaptation. We perform experience 
replay by sampling 100 examples from the 
memory and perform a gradient update after every 10,000 training 
steps, which gives us a 1\% replay rate.
\item \textsc{MbPA} \citep{mbpa}: an episodic memory model that uses stored examples for local adaptation without sparse experience replay. 
The original MbPA formulation has a trainable key network.
Our MbPA baseline uses a fixed key network since MbPA with
a trainable key network performs significantly worse.
\item \textsc{MbPA}$^{\text{rand}}_{\text{\arc}}$: an episodic memory model with randomly retrieved examples for local adaptation (no key network).
\item \textsc{MbPA}\arc{}: our episodic memory model described in \S{\ref{sec:model}}.
\item \textsc{MTL}: a multitask model trained on all datasets jointly, used as a performance upper bound.
\end{itemize}
%\dycomment{add another baseline if example forgetting works: random forgetting}

\subsection{Implementation Details}
We use a pretrained $\textsc{BERT}_{\textsc{BASE}}$ model \citep{bert}\footnote{\url{https://github.com/google-research/bert}}
as our example encoder and key network.
$\textsc{BERT}_{\textsc{BASE}}$ has 12 Transformer layers, 
12 self-attention heads, and 768 hidden dimensions (110M parameters in total).
% There are 110 million parameters in total in
% $\textsc{BERT}_{\textsc{BASE}}$.
%We update parameters of the example encoder but fix the key network.
We use the default BERT vocabulary in our experiments.

We use
Adam \citep{adam} as our optimizer.
We set dropout \citep{dropout} to 0.1 and
$\lambda$ in Eq. \ref{eq:local} to 0.001.
We set the base learning rate to $3e^{-5}$ (based on
preliminary experiments, in line with the suggested learning rate
for using \textsc{BERT}).
For text classification, we use a training batch of size 32.
For question answering, the batch size is 8.
The only hyperparameter that we tune is 
the local adaptation learning rate
$\in \{5e^{-3}, 1e^{-3}\}$.
We set the number of neighbors $K=32$ and the number of local 
adaptation steps $L=30$.
We show results with other $K$ and sensitivity 
to $L$ in \S{\ref{sec:analysis}}.

%\dycomment{cyp: add this A description of the computing infrastructure used.}
For each experiment, we use 4 Intel Skylake x86-64 CPUs at 2 GHz, 1 Nvidia Tesla V100 GPU, and 20 GB of RAM. 
% based on https://dm-xm2.corp.google.com/experiments/6024490 using 2 GCU on http://lubup6.prod.google.com:4901/

\subsection{Results}
\label{sec:results}
The models are trained in one pass on concatenated training sets, and
evaluated on the union of the test sets.
To ensure robustness of models to training dataset orderings, 
we evaluate on four different orderings (chosen randomly) for each task. As the multitask model has no inherent dataset ordering, 
we report results on four different shufflings of combined training examples. We show the exact orderings in Appendix~\ref{app:datasetorder}.
We tune the local adaptation learning rate using the 
first dataset ordering for each task and only run the best setting on the other orderings.

A main difference between these two tasks is that in text classification
the model acquires knowledge about new classes as training progresses (i.e.,
only a subset of the classes that corresponds to a particular dataset 
are seen at each training interval), whereas in question answering
the span predictor works similarly across datasets.

Table~\ref{tbl:results} provides a summary of our main results.
We report (macro-averaged) accuracy for 
classification and $F_1$ score\footnote{$F_1$ score is 
a standard question answering metric that measures $n$-grams overlap
between the predicted answer and the ground truth.} for question answering.
We provide complete per-dataset (non-averaged) results in Appendix~\ref{app:fullresults}.
Our results show that
\textsc{A-GEM} outperforms the standard encoder-decoder model \textsc{Enc-Dec},
although it is worse than \textsc{MbPA} on both tasks.
Local adaptation (\textsc{MbPA}) and sparse experience replay (\textsc{Replay}) help 
mitigate catastrophic forgetting
compared to \textsc{Enc-Dec}, but
a combination of them is needed to achieve the best performance (\textsc{MbPA}\arc).

%\dycomment{explain why a-gem is bad}

Our experiments also show that retrieving relevant examples from memory
is crucial to ensure that the local adaptation phase is useful.
Comparing the results from \textsc{MbPA}\arc{} and \textsc{MbPA}$^{\text{rand}}_{\text{\arc}}$,
we can see that the model that chooses neighbors randomly 
is significantly worse than the model that finds 
and uses similar examples for local adaptation.
We emphasize that having a fixed key network is 
crucial to prevent representation drift.
The original \textsc{MbPA} formulation that updates
the key network during training results in a model that
only performs slightly better 
than \textsc{MbPA}$^{\text{rand}}_{\text{\arc}}$ in our
preliminary experiments.
%far below the performance of \textsc{EpMem}.
Our results suggest that our best model 
can be improved further by choosing relevant examples 
for sparse experience replay as well. 
We leave investigations of such methods to future work.

% We also report results of a true multitask model as an 
% upper bound on achievable performance.
% This model is an encoder-decoder model 
% %(without any memory component)
% that sees all training examples in a multitask setting.
%(instead of our continual learning setup). 
% The multi-task model, which acts as an upper bound on achievable performance, 
% has been shown not to suffer from catastrophic forgetting \citep{glipaper}.
Comparing to the performance of the 
multitask model \textsc{MTL}---which is as an 
upper bound on achievable performance---we observe that 
there is still a gap between continual models
and the multitask model.\footnote{
Performance on each dataset 
with the multitask model is better than or comparable 
to a single dataset model that is trained only on that dataset.
Averaged performance of the multitask model across datasets on each task
is also better than single-dataset models.} 
\textsc{MbPA}\arc{} has the 
smallest performance gap.
For text classification, \textsc{MbPA}\arc{} outperforms 
single-dataset models in
terms of averaged performance (70.6 vs. 60.7), demonstrating 
the success of positive transfer.
For question answering, \textsc{MbPA}\arc{} still lags behind single
dataset models (62.0 vs. 66.0).
Note that the collection of single-dataset models have 
many more parameters since there is a different 
set of model parameters per dataset.
See Appendix~\ref{sec:singletask} for detailed results 
of multitask and single-dataset models.

\begin{figure}
\centering
\begin{subfigure}{.32\textwidth}
\includegraphics[width=.99\linewidth]{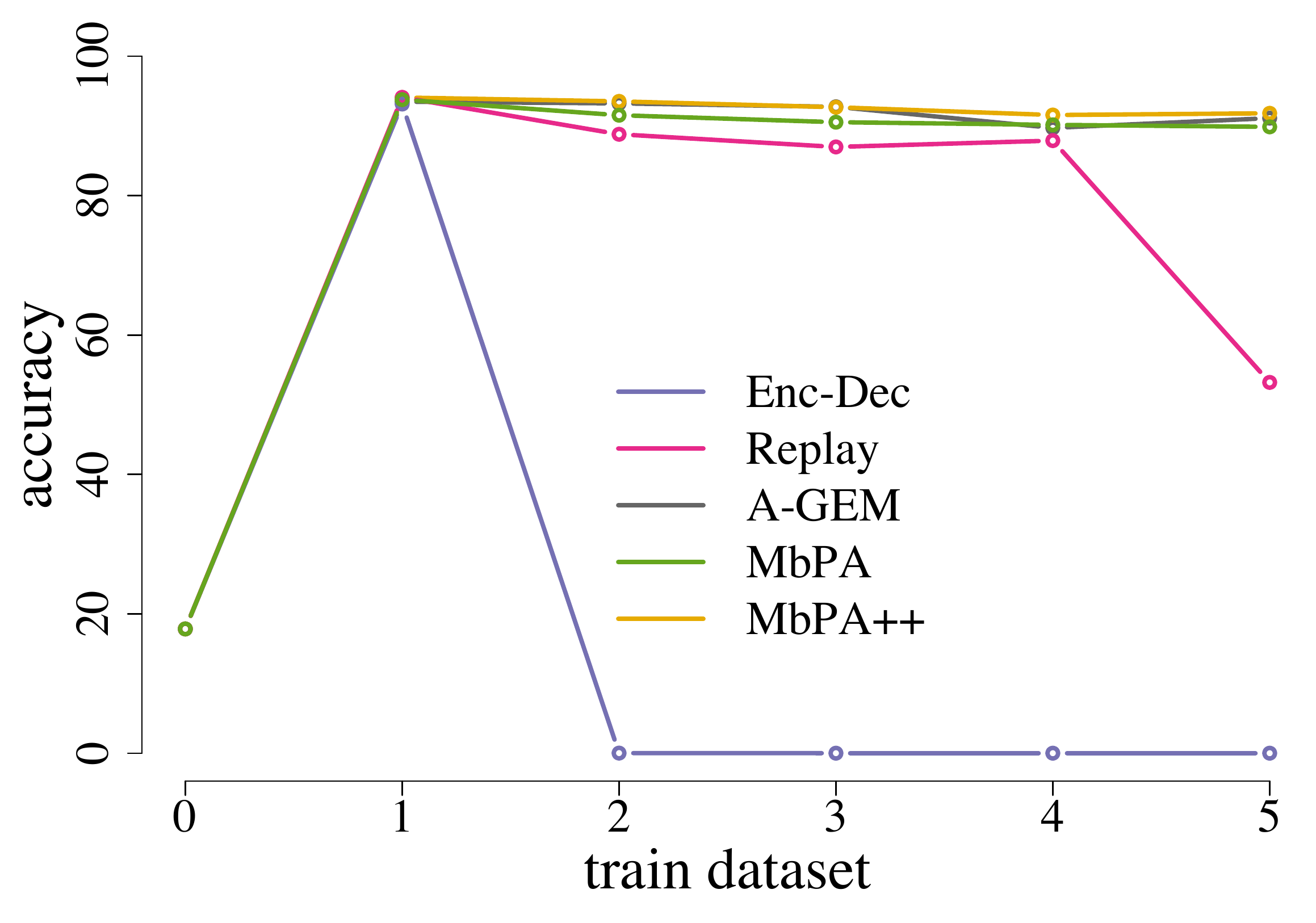}
\caption{Classification--AGNews}
\end{subfigure}
\begin{subfigure}{.32\textwidth}
\includegraphics[width=.99\linewidth]{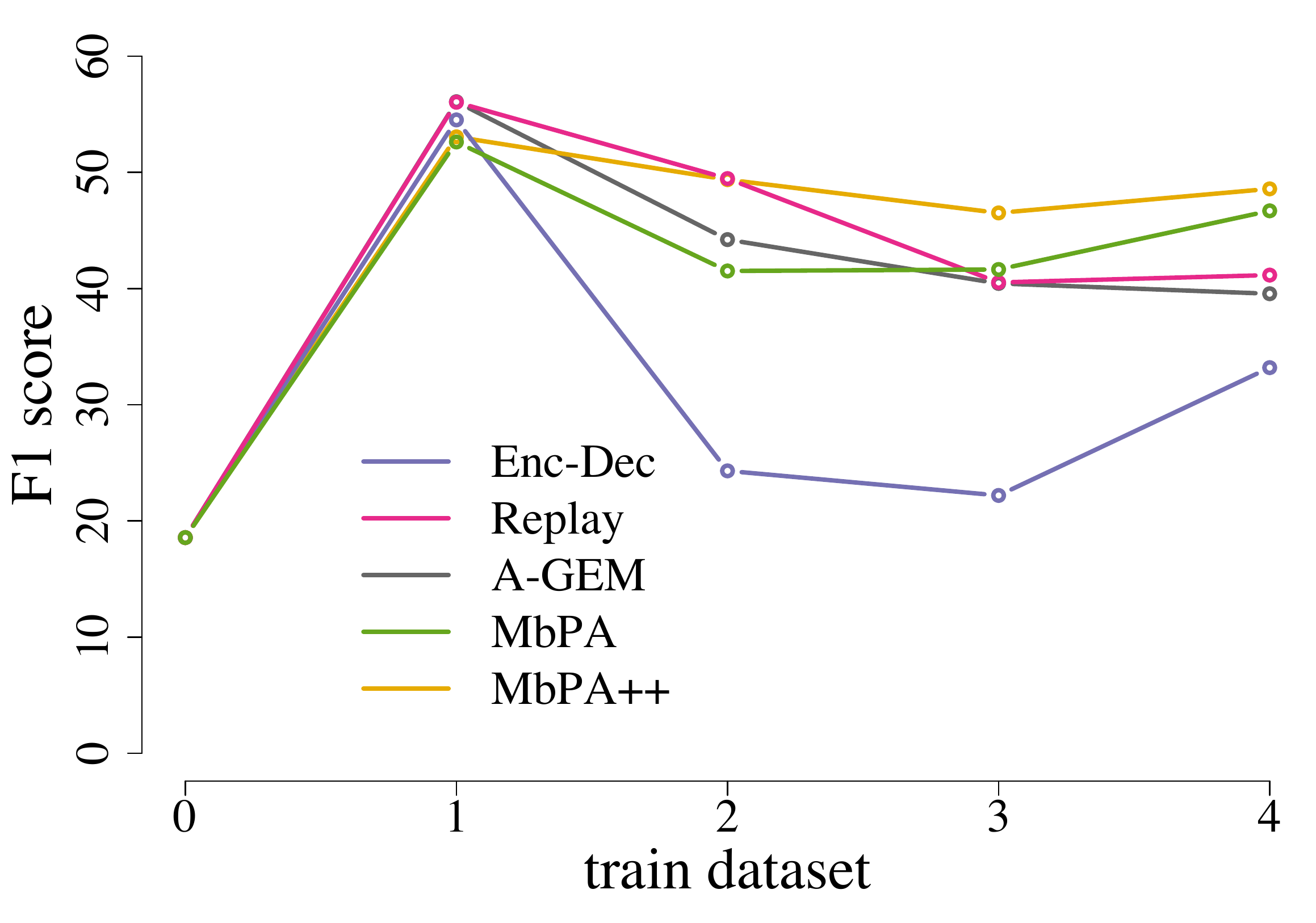}
\caption{QA--QuAC}
\end{subfigure}
\begin{subfigure}{.32\textwidth}
\includegraphics[width=.99\linewidth]{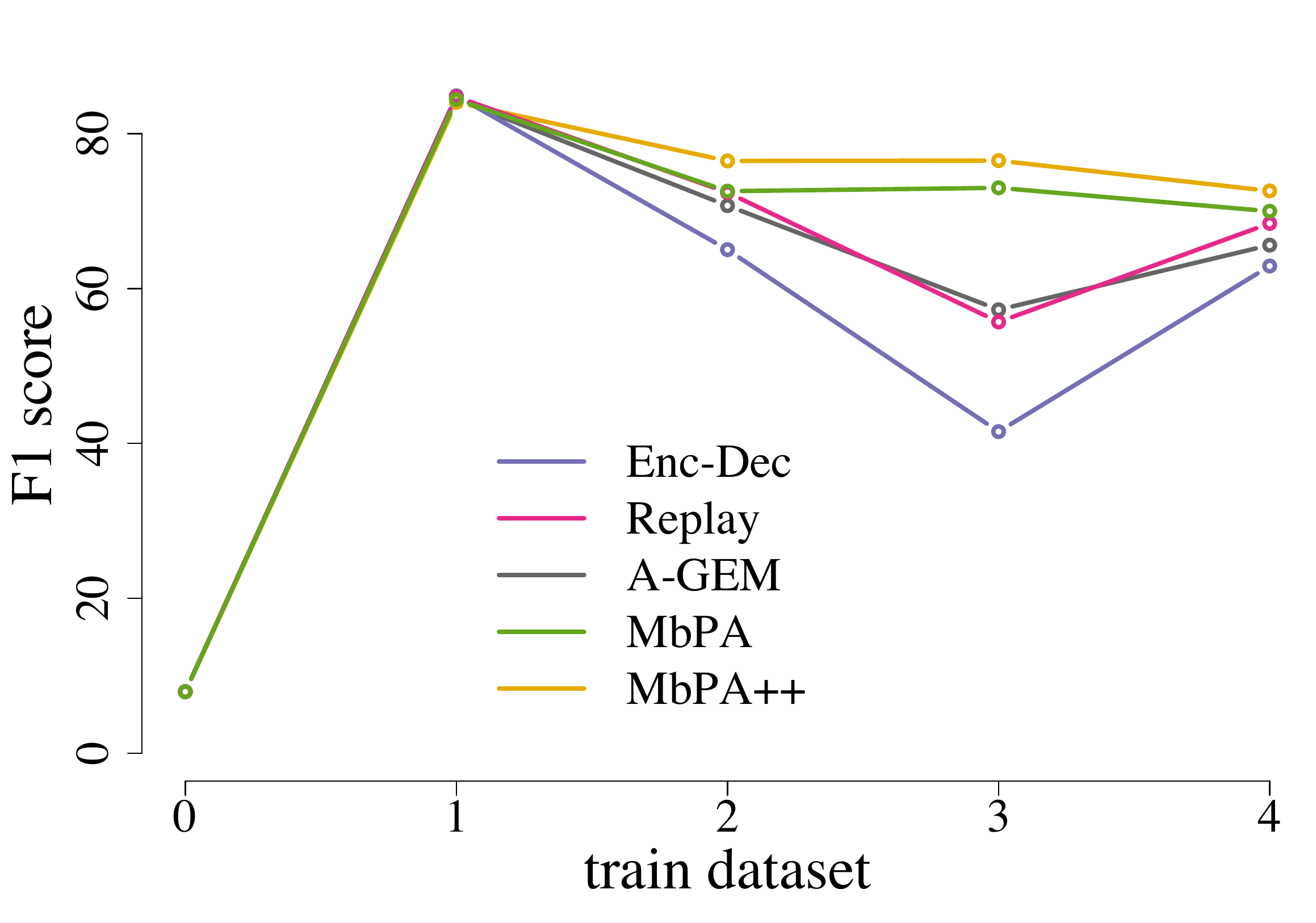}
\caption{QA--SQuAD}
\end{subfigure}
\vspace{-0.1cm}
\caption{Performance on test examples corresponding to the first dataset seen during training as training progresses.}
\vspace{-0.7cm}
\label{fig:lifetimeqa}
\end{figure}

%Table~\ref{tbl:results} provides a summary of results on all test datasets
%after seeing all training examples.
%We show results of 
%from one dataset order throughout the 
%lifetime of the models.
Figure~\ref{fig:lifetimeqa} shows 
$F_1$ score and accuracy of various models on 
the test set corresponding to the 
first dataset seen during training as the models are trained on more datasets.
The figure illustrates how well each model retains 
its previously acquired knowledge as it learns new knowledge.
We can see that \textsc{MbPA}\arc{} is consistently better
compared to other methods.

%\paragraph{Memory keys.}
%context+question and question only

\begin{table}[t]
%\dycomment{make table nice}
    \centering
    \caption{Summary of results on text classification (above) and question answering (below) using averaged accuracy and $F_1$ score respectively (see Appendix~\ref{app:datasetorder} for the dataset orderings).} \label{tbl:results}
    \begin{tabular}{lcccccc|c}
     \toprule
      \textbf{Order} & \textbf{\textsc{Enc-Dec}} & \textbf{\textsc{A-GEM}} & \textbf{\textsc{Replay}} & \textbf{\textsc{MbPA}} & \textbf{\textsc{MbPA}$^{\text{rand}}_{\text{\arc}}$} & \textbf{\textsc{MbPA}\arc} & \textbf{\textsc{MTL}}\\
      \midrule
    %   \multicolumn{8}{c}{Text classification}\\
    %   \midrule
      i & 14.8 & 70.6 & 67.2 & 68.9  & 59.4 & \textbf{70.8} & 73.7\\
      ii & 27.8 & 65.9 & 64.7 & 68.9 & 58.7 & \textbf{70.9} &73.2\\
      iii & 26.7& 67.5 & 64.7 & 68.8 & 57.1 & \textbf{70.2} & 73.7\\
      iv & 4.5 & 63.6 & 44.6 & 68.7 & 57.4& \textbf{70.7} & 73.7\\
      \midrule
      class.-avg. & 18.4 & 66.9 & 57.8 & 68.8 & 58.2 & \textbf{70.6} & 73.6\\
      \midrule
      \midrule
    %   \multicolumn{8}{c}{Question answering}\\
    %   \midrule
      i & 57.7 & 56.1 & 60.1 & 60.8 & 60.0 & \textbf{62.0} & 67.6 \\
      ii & 55.1 & 58.4 & 60.3 & 60.1 & 60.0 & \textbf{62.4} & 67.9\\
      iii & 41.6 & 52.4 & 58.8 & 58.9 & 58.8 & \textbf{61.4} & 67.9\\
      iv & 58.2 & 57.9 & 59.8 & 61.5 & 59.8 & \textbf{62.4} & 67.8 \\
      \midrule
      QA-avg. & 53.1 & 56.2 & 57.9 & 60.3 & 59.7 & \textbf{62.4} & 67.8\\
      \bottomrule
    \end{tabular}
    \vspace{-0.2cm}
\end{table}

\subsection{Analysis}
\label{sec:analysis}

\paragraph{Memory capacity.}
Our results in \S{\ref{sec:results}} assume that we can store all examples in memory (for all models, including the baselines).
We investigate variants of \textsc{MbPA}\arc{} that store only 50\% and 10\%
of training examples. We randomly decide whether
to write an example to memory or not (with probability 0.5 or 0.1).
We show the results in Table~\ref{tbl:memory}.
The results demonstrate that while the performance of the model 
degrades as the number of stored examples decreases, the model is still able
to maintain a reasonably high performance even with only 10\% memory
capacity of the full model.

\paragraph{Number of neighbors.}
We investigate the effect of the number of retrieved examples for local 
adaptation to the performance of the model in Table~\ref{tbl:neighbors}.
In both tasks, the model performs better as the number of neighbors
increases.\footnote{We are not able to obtain results for question answering with $K=64$ and $K=128$ due to out of memory issue (since the input text for question answering can be very long).}
Recall that the goal of the local adaptation phase is to
shape the output distribution of a test example to peak 
around relevant classes
(or spans) based on retrieved examples from the memory.
As a result, it is reasonable for the performance of the model
to increase with more neighbors (up to a limit) given a 
key network that can reliably compute similarities 
between the test example and stored examples in 
memory and a good adaptation method.

\begin{table}[]
    \begin{minipage}{.48\linewidth}
    % \vspace{-0.3cm
    \caption{Results with limited memory capacity.} 
    \label{tbl:memory}
    \centering
    %\resizebox{\textwidth}{!}
    \begin{tabular}{lccc}
     \toprule
       & 10\% & 50\% & 100\% \\
      \midrule
      class. & 67.6 & 70.3 & 70.6 \\
      QA & 61.5 & 61.6 & 62.0 \\
      \bottomrule
    \end{tabular}
    \vspace{-0.3cm}
    \end{minipage}%
    %\vspace{0.2mm}
    \hfill
    \begin{minipage}{.48\linewidth}
    % \vspace{-0.3cm}
    \centering
    \caption{Results for different \# of retrieved examples $K$.}
    \label{tbl:neighbors}
    \begin{tabular}{lccccc}
     \toprule
       & 8 & 16 & 32 & 64 & 128 \\
      \midrule
      class. & 68.4 & 69.3 & 70.6 & 71.3 & 71.6 \\
      QA & 60.2 & 60.8 & 62.0  & - & -\\
      \bottomrule
    \end{tabular}
    \vspace{-0.3cm}
    \end{minipage} 
    % \caption{Global caption}
\end{table}

\paragraph{Computational complexity.}
Training \textsc{MbPA}\arc{} takes as much time as training 
an encoder-decoder model
without an episodic memory module since experience replay is performed
sparsely (i.e., every 10,000 steps) with only 100 examples.
This cost is negligible in practice and we observe no significant 
difference in terms of
wall clock time to the vanilla encoder-decoder baseline.
\textsc{MbPA}\arc{} has a higher space complexity for
storing seen examples, which could be controlled
by limiting the memory capacity.

\begin{wrapfigure}{r}{0.45\textwidth}
\vspace{-0.7cm}
\begin{center}
    \includegraphics[scale=0.28]{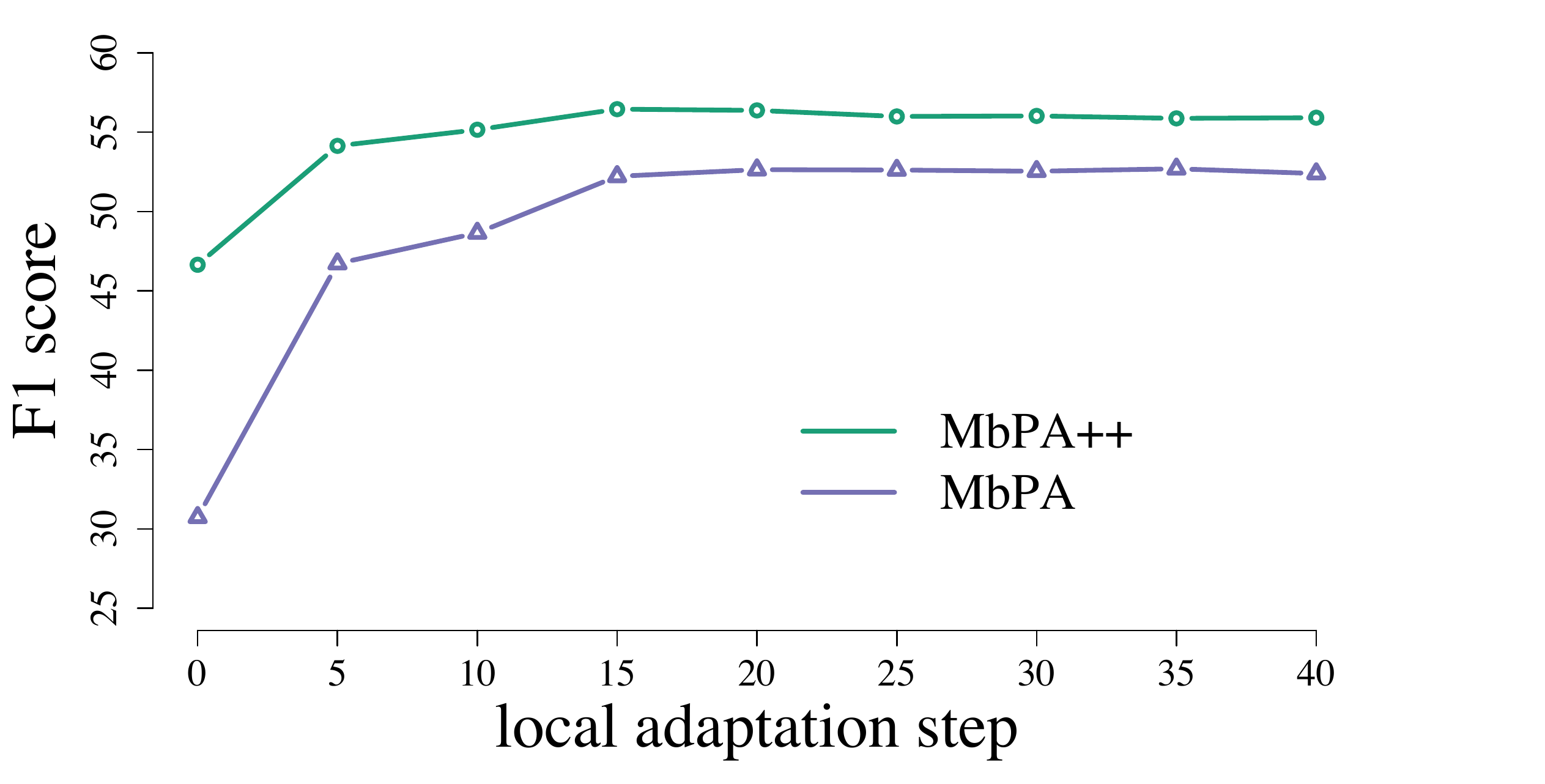}
\end{center}
\vspace{-0.5cm}
    \caption{$F_1$ scores for \textsc{MbPA}\arc{} and \textsc{MbPA} as the \# of local adaptation steps increases.} %We can also see that sparse experience replay help the \textsc{MbPA}\arc{} model to start from a better starting point and converge to a better final performance.}
    \label{fig:nsteps}
\vspace{-0.4cm}
\end{wrapfigure}

At inference time, \textsc{MbPA}\arc{} requires 
a local adaptation phase and is thus slower than methods
without local adaptation. This can be seen as a limitation of \textsc{MbPA}\arc{} (and \textsc{MbPA}).
One way to speed it up is to parallelize predictions across test examples,
since each prediction is independent of others.
We set the number of local adaptation steps $L=30$ in our experiments.
Figure~\ref{fig:nsteps} shows $L\approx15$ is needed to converge to an optimal performance.

Comparing \textsc{MBpA}\arc{} to other episodic memory models, 
\textsc{MBpA} has roughly 
the same time and space complexity
as \textsc{MBpA}\arc{}. \textsc{A-GEM}, on the other hand,
is faster at prediction time (no local adaptation), although at
training time it is slower due to extra projection steps and 
uses more memory since it needs to store
two sets of gradients (one from the current batch, and one
from samples from the memory). We find that this 
cost is not negligible when 
using a large encoder such as \textsc{BERT}.

\paragraph{Analysis of retrieved examples.} In
Appendix~\ref{app:retrievedexamples},
we show (i) examples of retrieved neighbors from 
our episodic memory model, (iii) examples
where local adaptation helps, and (iii) examples
that are difficult to retrieve.
We observe that the model is able to retrieve examples
that are both syntactically and semantically related
to a given query derived from a test example, especially
when the query is not too short and relevant
examples in the memory are phrased in a similar way.
%For example, in question answering, we find that using 
%only the question as the key produces samples that ask for
%very similar things and which should 
%thus be particularly beneficial for local adaptation.
%\dycomment{cyp add here}
%check rapid adaptation, incremental learning

\section{Conclusion}
We introduced a lifelong language learning setup and presented an
episodic memory model that performs sparse experience 
replay and local adaptation
to continuously learn and reuse previously acquired knowledge.
Our experiments
demonstrate that our proposed method mitigates 
catastrophic forgetting and outperforms baseline methods
on text classification and question answering.

\subsubsection*{Acknowledgements}
We thank Gabor Melis and the three anonymous reviewers for helpful feedback on an earlier draft of this paper.

\bibliography{multitask_memory}
\bibliographystyle{icml2019}

\appendix

\section{Dataset Order}
\label{app:datasetorder}
We use the following dataset orders (chosen randomly) for text classification: 
\begin{enumerate}[topsep=0pt, partopsep=0pt,itemsep=1pt,parsep=1pt,label=(\roman{*})]
\item Yelp $\rightarrow{}$ AGNews $\rightarrow{}$ DBPedia $\rightarrow{}$ Amazon $\rightarrow{}$ Yahoo.
\item DBPedia $\rightarrow{}$ Yahoo $\rightarrow{}$ AGNews $\rightarrow{}$ Amazon $\rightarrow{}$ Yelp.
\item Yelp $\rightarrow{}$ Yahoo $\rightarrow{}$ Amazon $\rightarrow{}$ DBpedia $\rightarrow{}$ AGNews.
\item AGNews $\rightarrow{}$ Yelp $\rightarrow{}$ Amazon $\rightarrow{}$ Yahoo $\rightarrow{}$ DBpedia.
\end{enumerate}

For question answering, the orders are:
\begin{enumerate}[topsep=0pt, partopsep=0pt,itemsep=1pt,parsep=1pt,label=(\roman{*})]
\item QuAC $\rightarrow{}$ TrWeb $\rightarrow{}$ TrWik $\rightarrow$ SQuAD.
\item SQuAD $\rightarrow{}$ TrWik $\rightarrow{}$ QuAC $\rightarrow$ TrWeb.
\item TrWeb $\rightarrow{}$ TrWik $\rightarrow{}$ SQuAD $\rightarrow$ QuAC.
\item TrWik $\rightarrow{}$ QuAC $\rightarrow{}$ TrWeb $\rightarrow$ SQuAD.
\end{enumerate}

\section{Full Results}
\label{app:fullresults}

We show per-dataset breakdown of results in Table~\ref{tbl:results} in Table~\ref{tbl:fullresultstextcat} and Table~\ref{tbl:fullresultsqa} for text classification and question answering respectively.

\begin{table}[ht]
    \caption{Per-dataset results on text classification for each ordering and model.}
    \label{tbl:fullresultstextcat}
\renewcommand{\arraystretch}{1.1}
\renewcommand{\tabcolsep}{4mm}
    \centering
    \begin{tabular}{ccccccc}
     \toprule
      \multirow{2}{*}{\textbf{Order}} & \multirow{2}{*}{\textbf{Model}} &
      \multicolumn{5}{c}{\textbf{Dataset}} \\
      \cmidrule(l){3-7}
      & & \textbf{1} & \textbf{2} & \textbf{3} & \textbf{4} & \textbf{5}\\
      \midrule
      \multirow{6}{*}{i} & \textsc{Enc-Dec} & 1.1 & 0.0 & 0.0 & 4.3 & 68.7 \\
      & \textsc{A-GEM} & 42.5 & 89.8 & 96.0 & 56.8 & 68.2 \\
      & \textsc{Replay} & 38.2 & 83.9 & 95.4 & 50.3 & 67.9 \\
      & \textsc{MbPA} & 42.0 & 90.4 & 96.1 & 52.0 & 63.9\\
      & \textsc{MbPA}$^{\text{rand}}_{\text{\arc}}$ & 35.2 & 80.4 & 88.2 & 45.9 & 47.2 \\
      & \textsc{MbPA}\arc & 45.7 & 91.6 & 96.3 & 54.6 & 65.6\\
      \midrule
      \multirow{6}{*}{ii} & \textsc{Enc-Dec} & 0.0 & 0.0 & 3.1 & 57.9 & 48.9 \\
      & \textsc{A-GEM} & 80.1 & 50.3 & 91.3 & 57.3 & 50.6 \\
      & \textsc{Replay} & 75.0 & 53.7 & 86.0 & 58.1 & 50.7 \\
      & \textsc{MbPA} & 96.0 & 58.4 & 89.0 & 54.4 & 46.6\\
      & \textsc{MbPA}$^{\text{rand}}_{\text{\arc}}$ & 82.0 & 41.7 & 81.9 & 47.1 & 40.8 \\
      & \textsc{MbPA}\arc & 95.8 & 63.1 & 92.2 & 55.7 & 47.7 \\
      \midrule
      \multirow{6}{*}{iii} & \textsc{Enc-Dec} & 0.0 & 0.0 & 1.3 & 11.4 & 93.9 \\
      & \textsc{A-GEM} & 41.1 & 55.0 & 54.6 & 93.3 & 93.6 \\
      & \textsc{Replay} & 23.6 & 36.8 & 25.0 & 94.5 & 93.8 \\
      & \textsc{MbPA} & 43.3 & 60.9 & 51.6 & 95.8 & 92.5 \\
      & \textsc{MbPA}$^{\text{rand}}_{\text{\arc}}$ & 35.2 & 33.6 & 42.1 & 92.3 & 82.3 \\
      & \textsc{MbPA}\arc & 44.3 & 62.7 & 54.4 & 96.2 & 93.4 \\
      \midrule
      \multirow{6}{*}{iv} & \textsc{Enc-Dec} & 0.0 & 0.0 & 0.0 & 14.1 & 8.1 \\
      & \textsc{A-GEM} & 90.8 & 44.9 & 60.2 & 65.4 & 56.9 \\
      & \textsc{Replay} & 70.4 & 33.2 & 39.8 & 46.1 & 33.4 \\
      & \textsc{MbPA} & 89.9 & 42.9 & 52.6 & 62.9 & 95.1 \\
      & \textsc{MbPA}$^{\text{rand}}_{\text{\arc}}$ & 78.4 & 37.7 & 45.8 & 42.4 & 82.9 \\
      & \textsc{MbPA}\arc & 91.8 & 44.9 & 55.7 & 65.3 & 95.8 \\
      \bottomrule
    \end{tabular}
\end{table}

\begin{table}[ht]
    \caption{Per-dataset results on question answering for each ordering and model.}
    \label{tbl:fullresultsqa}
\renewcommand{\arraystretch}{1.1}
\renewcommand{\tabcolsep}{4mm}
    \centering
    \begin{tabular}{cccccc}
     \toprule
      \multirow{2}{*}{\textbf{Order}} & \multirow{2}{*}{\textbf{Model}} &
      \multicolumn{4}{c}{\textbf{Dataset}} \\
      \cmidrule(r){3-6}
      & & \textbf{1} & \textbf{2} & \textbf{3} & \textbf{4}\\
      \midrule
      \multirow{6}{*}{i} & \textsc{Enc-Dec} & 34.1 & 54.2 & 56.0 & 85.5 \\
      & \textsc{A-GEM} & 36.7 & 51.8 & 53.4 & 82.5 \\
      & \textsc{Replay} & 40.9 & 56.7 & 57.2 & 85.8 \\
      & \textsc{MbPA} & 45.6 & 56.1 & 57.9 & 83.4 \\
      & \textsc{MbPA}$^{\text{rand}}_{\text{\arc}}$ & 41.5 & 56.7 & 57.2 & 85.8 \\
      & \textsc{MbPA}\arc & 47.2 & 57.7 & 58.9 & 84.3 \\
      \midrule
      \multirow{6}{*}{ii} & \textsc{Enc-Dec} & 61.9 & 64.2 & 29.3 & 65.0 \\
      & \textsc{A-GEM} & 64.2 & 62.5 & 43.4 & 63.5 \\
      & \textsc{Replay} & 67.0 & 64.1 & 44.9 & 65.2 \\
      & \textsc{MbPA} & 69.9 & 63.4 & 43.6 & 63.3 \\
      & \textsc{MbPA}$^{\text{rand}}_{\text{\arc}}$ & 67.5 & 62.5 & 46.5 & 63.7 \\
      & \textsc{MbPA}\arc & 72.6 & 63.4 & 50.5 & 63.0 \\
      \midrule
      \multirow{6}{*}{iii} & \textsc{Enc-Dec} & 30.7 & 31.2 & 45.6 & 58.7 \\
      & \textsc{A-GEM} & 47.6 & 47.0 & 57.4 & 57.4 \\
      & \textsc{Replay} & 46.6 & 45.4 & 53.9 & 58.3 \\
      & \textsc{MbPA} & 52.5 & 54.6 & 74.5 & 54.3 \\
      & \textsc{MbPA}$^{\text{rand}}_{\text{\arc}}$ & 54.1 & 54.3 & 71.1 & 55.9 \\
      & \textsc{MbPA}\arc & 56.0 & 56.8 & 78.0 & 54.9 \\
      \midrule
      \multirow{6}{*}{iv} & \textsc{Enc-Dec} & 55.5 & 37.1 & 54.8 & 85.4 \\ 
      & \textsc{A-GEM} & 54.8 & 38.8 & 53.4 & 84.7 \\
      & \textsc{Replay} & 56.9 & 41.8 & 56.4 & 86.1 \\
      & \textsc{MbPA} & 58.0 & 47.2 & 57.4 & 83.3 \\
      & \textsc{MbPA}$^{\text{rand}}_{\text{\arc}}$ & 55.5 & 43.0 & 54.6 & 85.9 \\
      & \textsc{MbPA}\arc & 59.0 & 48.7 & 58.1 & 83.6 \\
      \bottomrule
    \end{tabular}
\end{table}

\section{Single Dataset Models}
\label{sec:singletask}
We show results of a single dataset model that is only trained on
a particular dataset in Table~\ref{tbl:singletask}.

\begin{table}[ht]
\centering
\caption{Performance of a standard encoder-decoder model on each dataset in our experiments. We report accuracy for text classification and $F_1$ score for question answering. We also show results from a multitask model for comparisons.} \label{tbl:singletask}
\renewcommand{\arraystretch}{1.1}
\renewcommand{\tabcolsep}{4mm}
\begin{tabular}{clcc}
\toprule
\textbf{Task} & \textbf{Dataset} & \textbf{Single Model} & \textbf{Multitask}\\
\midrule
\multirow{6}{*}{Text Classification} & AGNews & 93.8 & 94.0\\
& Yelp & 50.7 & 50.3\\ 
& Amazon & 60.1 & 58.8\\
& Yahoo & 68.6 & 67.1\\
& DBPedia & 30.5 & 95.9\\
\midrule
& Average & 60.7 & 73.2\\
\midrule
\multirow{5}{*}{Question Answering} & QuAC & 54.3 & 56.4\\
& SQuAD & 86.1 & 85.7\\
& Trivia Wikipedia & 62.3 & 64.0\\
& Trivia Web & 62.4 & 64.4\\
\midrule
& Average & 66.0 & 67.6\\
\bottomrule
\end{tabular}
\end{table}

\section{Analysis of Retrieved Examples}
\label{app:retrievedexamples}
We analyze retrieved examples to better
understand how our model uses its episodic memory.

\paragraph{Examples of retrieved neighbors.}
We show examples of retrieved neighbors from memory 
given a test query in Table~\ref{tbl:retrievedexamples}.
We observe that the model is generally able to retrieve
relevant examples from the memory. 
In question answering, nearest neighbors 
tend to be examples that are both syntactically
and semantically related.
In text classification, they tend to be
articles that discuss the a similar topic.

\paragraph{Examples where local adaptation helps.}
%For question answering, we show some test examples where local adaptation helps in Table~\ref{tbl:help}.
%For text classification, we found that local adaptation is almost always necessary to correctly classify test examples from datasets that are encountered at the beginning of training. Without local adaptation, (as also indicated by the results of \textsc{Enc-Dec} and \textsc{Replay} models).
%
In Table~\ref{tbl:help}, we show two test examples where our model answers incorrectly before local adaptation, but correctly after. In the first case, we can see that training examples retrieved from memory are thematically related to the test example. In the second case, since the query is shorter, retrieved training examples tend to be more syntactically related. Although we only show the two nearest neighbors for each query here, our analysis provides an insight on ways our model uses its memory to improve predictions.

%When local adaptation does not help, We found that examples
\paragraph{Relevant examples that are difficult to retrieve.}
In Table~\ref{tbl:relevant}, we show two relevant training examples (as judged by humans) that are difficult to retrieve by the model (they are not in the 1,000 nearest neighbors) for the query \texttt{what was the name of bohemond s nephew}. The two relevant training examples ask about the nephew of a person, which is relevant for the given query.
However, since they are phrased differently to the query, they are far in the embedding space, which is why a nearest neighbor method fails to retrieve these training examples.
Our analysis shows that a better embedding and/or retrieval method can potentially improve the performance of our model.
%The model retrieves questions formulated similarly but asking about unrelated properties of unrelated people, instead of retrieving thematically relevant questions formulated differently.
% \begin{table*}[h]
% \centering
% \caption{template for tables.}
% \label{tbl:relevant}
% \begin{tabular}{p{0.5\linewidth}}
% \toprule
% % test example 62, train example 27966
% \textbf{Query:} what was name of bohemond s nephew  \\ \midrule
% \textbf{First 3 nearest neighbors:} \\
% (49.73) what was isambard brunel s occuption \\
% (52.54) what was jef raskin s profession \\
% (56.00) what was richard scudamore s job \\
% \midrule
% \textbf{Relevant examples not retrieved:} \\
% (87.88) who was nephew of leopold \\
% (103.96) who is nephew of buda king casimer iii great \\
% \bottomrule
% \end{tabular}
% \end{table*}

\begin{table*}[h]
\caption{Examples of queries and retrieved nearest neighbors for question answering (top) and text classification (truncated, bottom). We also show the corresponding Euclidean distances in parentheses.}
\label{tbl:retrievedexamples}
\begin{tabular}{p{\linewidth}}
\toprule
\textbf{Query: in what country is normandy located}  \\ \midrule
(17.48) in what area of france is calais located \\
(20.37) in what country is st john s located \\
(22.76) in what country is spoleto located \\
(23.12) in what part of africa is congo located \\
(23.83) on what island is palermo located \\
\bottomrule
\vspace{0.1cm}
\end{tabular}

\begin{tabular}{p{\linewidth}}
\toprule
  \textbf{Query: fears for t n pension after talks unions representing workers at turner newall say they are disappointed after talks with stricken parent firm federal mogul}  \\ \midrule
(37.32) union anger at sa telecoms deal south african unions describe as disgraceful use of public money to buy telecoms shares for former government officials\\
(47.60) us hurting anti mine campaign anti landmine activists meeting in nairobi say us is setting bad example by not joining worldwide ban\\
(49.03) woolworths ordered to extend alh takeover deadline independent takeovers panel has headed off woolworths 39 attempts to force resolution in takeover battle for liquor retailer australian leisure and hospitality alh\\
(50.42) price hike for business broadband small net firms warn they could be hit hard by bt s decision to raise prices for business broadband\\
(51.08) job fears as deutsche culls chiefs deutsche bank is streamlining management of its investment banking arm raising fears that jobs may be lost in city german bank is reducing number of executives running its investment banking\\
\bottomrule
\end{tabular}
\end{table*}

\begin{table*}[h]
\centering
\vspace{-0.2cm}
\caption{Two examples where local adaptation  helps.}
\label{tbl:help}
\begin{tabular}{p{.9\linewidth}}
\toprule
% experiment_id = 6005108
% wid = 31
% eval_step = 0
\textbf{Context:} david niven ( actor ) - pics , videos , dating , \& news david niven male born mar 1 , 1910 james david graham niven , known professionally as david niven , was an english actor and novelist [\dots]\\ \midrule
\textbf{Query:} in 1959 , for which film did david niven win his only academy award ?\\ \midrule
\textbf{First two training examples retrieved from memory (2 nearest neighbors):} \\
in which of her films did shirley temple sing animal crackers in my soup ? \\
in 1968 , which american artist was shot and wounded by valerie solanis , an actress in one of his films ? \\
%in what year did schwarzenegger play jayne mansfield ' s husband in a film ? \\
\midrule
\midrule
% experiment_id = 6005108
% wid = 29
% eval_step = 0
\textbf{Context:} dj kool herc developed the style that was the blueprint for hip hop music . herc used the record to focus on a short , heavily percussive part in it : the " break " . [\dots]\\ \midrule
\textbf{Query:} what was the break ?\\ \midrule
\textbf{First two training examples retrieved from memory (2 nearest neighbors):} \\
what was the result ? \\
what was the aftermath ? \\
%what was the music ? \\
\bottomrule
\end{tabular}
\vspace{-0.2cm}
\end{table*}

\begin{table*}[h]
\vspace{-0.2cm}
\centering
\caption{Relevant examples that are difficult to retrieve from memory.}
\label{tbl:relevant}
\begin{tabular}{p{0.9\linewidth}l}
\toprule
% test example 62, train example 27966
\textbf{Query:} what was the name of bohemond s nephew  \\ \midrule
%\textbf{2 nearest neighbors:} & 
\textbf{Relevant examples not retrieved (Euclidean distances to the query in parentheses):} \\
%(49.73) what was isambard brunel s occupation 
(87.88) who was the nephew of leopold \\
%(52.54) what was jef raskin s profession & 
(103.96) who is the nephew of buda king casimer iii the great \\
% (56.00) what was richard scudamore s job & \\
\bottomrule
\end{tabular}
\vspace{-0.1cm}
\end{table*}

\end{document}